\documentclass[letterpaper]{article} 
\usepackage{aaai25}  
\usepackage{times}  
\usepackage{helvet}  
\usepackage{courier}  
\usepackage[hyphens]{url}  
\usepackage{graphicx} 
\usepackage{natbib}  
\usepackage{caption} 
\usepackage{algorithm}
\usepackage{algorithmic}
\usepackage{amsmath}
\usepackage{multirow}
\usepackage{amsfonts}

\usepackage{newfloat}
\usepackage{listings}
\DeclareCaptionStyle{ruled}{labelfont=normalfont,labelsep=colon,strut=off} 
\floatstyle{ruled}
\newfloat{listing}{tb}{lst}{}
\floatname{listing}{Listing}
\title{Guaranteeing Out-Of-Distribution Detection in Deep RL via Transition Estimation}
\author{
    Mohit Prashant, Arvind Easwaran, Suman Das, Michael Yuhas\\
}
\affiliations{
    
    Nanyang Technological University\\
    \{ mohit010@e., arvinde@, suman.das@, michaelj004@e. \} ntu.edu.sg\\
}

\usepackage{bibentry}

\begin{document}

\maketitle

\begin{abstract}
An issue concerning the use of deep reinforcement learning (RL) agents is whether they can be trusted to perform reliably when deployed, as training environments may not reflect real-life environments. Anticipating instances outside their training scope, learning-enabled systems are often equipped with out-of-distribution (OOD) detectors that alert when a trained system encounters a state it does not recognize or in which it exhibits uncertainty. There exists limited work conducted on the problem of OOD detection within RL, with prior studies being unable to achieve a consensus on the definition of OOD execution within the context of RL. By framing our problem using a Markov Decision Process, we assume there is a transition distribution mapping each state-action pair to another state with some probability. Based on this, we consider the following definition of OOD execution within RL: A transition is OOD if its probability during real-life deployment differs from the transition distribution encountered during training. As such, we utilize conditional variational autoencoders (CVAE) to approximate the transition dynamics of the training environment and implement a conformity-based detector using reconstruction loss that is able to guarantee OOD detection with a pre-determined confidence level. We evaluate our detector by adapting existing benchmarks and compare it with existing OOD detection models for RL.
\end{abstract}


\section{Introduction}

Ideally, an RL agent is trained to synthesize a policy based on a training environment that closely resembles its real-life use case. However, rare events/states may be encountered during deployment that are outside of the distribution of data encountered during the training process. Furthermore, due to rarity, sufficient training using exploration (i.e. without a generative simulator) is insufficient to guarantee the outcome of policy execution over these states. We refer to the execution of a learned policy over an unfamiliar/uncertain environment as out-of-distribution (OOD) with respect to the data encountered during the training process \cite{uncer1}. If a learned policy is applied to an environment that is significantly different to the training environment, it is expected that the policy will yield unpredictable outcomes \cite{OOD5}. As it is desirable to detect uncertainty to supplement training (or avoid completely in safety-critical settings), it is necessary to identify OOD execution. 


The absence of a concrete definition for OOD execution within the RL context is due to a number of reasons. Though they contribute to a difference between the training environment and the real-life environment, there are many potential causes of uncertainty in policy execution. Prior works identify some causes of OOD execution as the following: environment stochasticity \cite{uncer1}; perturbation to observed data \cite{OOD2}; encountering unfamiliar states during execution \cite{uncer5}; domain shifts and shifts in environmental factors \cite{OOD4, haider10}. 


Due to the number of independent underlying causes of OOD execution identified within the RL context by prior studies and the absence of a comprehensive definition for OOD detection, a limitation of existing studies is that they are unable to comprehensively address this issue. For instance, the methods of quantifying value uncertainty are able to address environment stochasticity, but are unable to detect perturbations as OOD \cite{uncer5}. Similarly, the detection of transition perturbations proposed to address domain shifts is unable to identify unfamiliar states as a cause of OOD execution \cite{OOD2}. 


\subsubsection{Contribution: Defining OOD Execution}

We propose two possible types of OOD execution: \textbf{(1)} \textit{state-based OOD execution}, when an agent reaches a state which was insufficiently encountered during training; and \textbf{(2)} \textit{environment-based OOD execution}, when an agent is deployed in an environment with a transition function that differs from that of the training environment. In doing so, we build upon the definition introduced in \cite{OOD2}.


\subsubsection{Contribution: Guaranteeing OOD Detection} 
We characterize OOD execution by learning the transition dynamics of the underlying environment using a variational autoencoder that is conditioned upon the occurring transition \cite{CVAE}. We apply a conformal prediction method upon the reconstruction error, also conditioned on the transition, which allows us to \emph{guarantee a true positive rate for OOD detection}. We demonstrate the efficacy of our technique in detecting OOD execution by utilizing benchmarks previously established by \cite{OOD5,bench1}. We demonstrate up to a $17\%$ improvement in detection from previous methods.




\section{Preliminaries}
\subsection{Markov Decision Processes}
We model our learning problem as a continuous Markov Decision Process (MDP) characterized by the parameters $(S, A, T, R, \gamma)$: $S$ is the state-space within which agents can occupy states; $A$ is the action-space, describing the actions agents can take to transition between states; $T : S \times S \times A \rightarrow [0, 1]$ is a function that determines the transition probability to a state given a state-action; $R : S \times A \rightarrow \mathbb{R}$ is the expected reward from executing a state-action; $\gamma \in (0, 1)$ is the discount factor for future rewards \cite{markov}. As the learning problem is model-free, note that $T$ and $R$ are unknown to the learner within this study. The aim of an RL algorithm applied to this MDP is to learn a state-wise policy that maximizes the expected cumulative discounted reward (value), $V$. Let $s_t \in S, a_t \in A$ represent the state and action at timestep $t$.

\begin{equation} \label{valuedef}
\small
    V(s_t) := \max_{a_t \in A} \left[ R(s_t, a_t) + \gamma \sum_{s_{t+1} \in S} V(s_{t+1}) T(s_{t+1} | s_t, a_t) \right]
\end{equation}

\subsection{Uncertainty in Learning} \label{sec2.1}
Uncertainty within learning is a measure of the randomness within the inferences drawn. By assuming there is an underlying probability distribution that characterizes randomness, prior studies have estimated uncertainty using properties like variance and entropy \cite{uncer5,uncer12}.

\subsubsection{Epistemic Uncertainty}
Epistemic uncertainty is randomness that is caused by a lack of information. That is, the randomness in the learner's inference is due to not having observed enough data. Using the setting of an MDP to contextualize epistemic uncertainty, RL algorithms sample from transition and reward functions to learn a policy that maximizes the expected state-wise value, allowing for the determination of the best possible action with high confidence \cite{offline1, offline2}.

\subsubsection{Aleatoric Uncertainty}
As opposed to epistemic uncertainty, which is a consequence of lacking information, aleatoric uncertainty is randomness that is inherent to the environment. As such, obtaining more information does not reduce the impact of aleatoric uncertainty on the inference being made. Within an MDP setting, prior studies model this by adding an element of stochasticity to transitions and perturbing the outcome of the inference \cite{bench1,OOD2}.

\subsection{Conditional Variational Autoencoders} 
Conditional Variational Autoencoders (CVAE) are a class of generative DNNs based on the distributional encode-decode approach of Variational Autoencoders (VAE) \cite{VAE, CVAE}. For an observation, $x$, that is conditioned on a label, $y$, the model assumes the existence of an underlying latent variable, $z$, that maps to the observation space and is distributed by an intractable prior, $\mathcal{P}(x|y)$, and a similarly intractable posterior, $\mathcal{P}(z|x, y)$. The aim of the encoder and decoder are to compute the posterior and likelihood using Gaussian approximations, $\mathcal{Q}_\theta (z|x, y)$ and $\mathcal{P}_\theta (x|z, y)$, respectively. Intuitively, the encoding and decoding takes place with respect to the label, with each label mapping to a different distribution of observations. The likelihood of observations being sampled and reconstructed accurately, for a label, depends on being seen frequently during training with respect to the label. Our study makes use of this property in constructing an OOD detector.

\subsection{Inductive Conformal Prediction}
Inductive conformal prediction (ICP) is a classification technique that relies on the degree to which future instances conform with known data and accordingly issues a guarantee on the confidence of the prediction \cite{ICP}. Let the set of all possible observations be denoted $X := \{x_a, x_b ...\}$ and the set of all potential labels be denoted $Y := \{ y_a, y_b ...\}$. Let a finite set of observations be denoted using $X_* := \{ x_1, x_2... x_n\}$, and a corresponding set of labels for each observation denoted using $Y_* := \{ y_1, y_2... y_n\}$, $\forall (x_i, y_i) \in (X_*, Y_*), (x_i, y_i) \in X \times Y$. Assuming there exists a function, $\Delta : X \times Y \rightarrow \mathbb{R}$, which assigns a score to any observation-label pair, there exists a set of corresponding scores, $C := \{ \Delta(x_1, y_1), \Delta(x_2, y_2) ... \Delta(x_n, y_n)\}$, for the elements of the calibration set $(X_*, Y_*)$. Through sorting $C$ in ascending order, for a significance value $\delta \in (0, 1)$, a threshold score is obtained from $\delta \times n$-th element of the sorted $C$; let this value be denoted $C_\delta$. Subsequently, to predict the label of an unlabeled observation, $x$, ICP assigns a set prediction, $\Gamma$, to the observation against all potential labels.

\begin{equation} \label{eq1}
    \Gamma (x) := \left\{ y \in Y | \Delta(x, y) > C_\delta  \right\}  
\end{equation}

\noindent The set is a prediction of all potential labels that can be assigned to the observation given their score conforms with the scores established in $C$. Assuming $C$ is sufficiently large and representative of the distribution of scores encountered during training, all predictions are made with a $1-\delta$ confidence. Conversely, a $\phi$ prediction is an indication of an observation being OOD with the distribution of the calibration set.


\section{Related Works} \label{sec3}
\subsection{Uncertainty Measurement}

Uncertainty measurement is a field of study that has overlapping characteristics with OOD detection, with several OOD detectors utilizing uncertainty when making predictions \cite{uncer1,uncer3}. The objective of these studies is to parameterize the uncertainty experienced by the RL model in various aspects of learning. These include: measuring the variance in value with respect to the state-space \cite{uncer6,uncer5}; measuring the entropy of a policy in determining actions from a state \cite{uncer4,uncer7}; and measuring the entropy of the trajectory from a state, and hence the uncertainty of the transition dynamics \cite{uncer8}. It should be noted that the body of work of OOD detection within the context of offline-RL also attempts to identify unfamiliar states using forms of uncertainty measurement \cite{offline3, uncer14, uncer6, offline2}. This is a non-exhaustive list on uncertainty within RL and \cite{uncer9} provides a review on some of the specific techniques used within deep RL. However, the primary motivations behind measuring uncertainty are to either encourage exploration or to avoid uncertain scenarios, the latter of which is closer to OOD detection. We focus our review on works that aim to avoid uncertain scenarios. Techniques involve: estimating uncertainty empirically \cite{uncer12,uncer5,uncer15}; as well as incorporating uncertainty within the policy process by de-incentivizing the agent from taking uncertain actions \cite{uncer11,uncer4,uncer18,uncer16,uncer13,uncer17}. We posit that uncertainty estimation is an integral part of OOD detection, with uncertain states and trajectories being OOD. However, they are unable to account for all forms of OOD execution.

\subsection{Defining OOD Execution}
A definition of OOD execution is one where the training environment is a poor approximation of the real-life environment due to domain shift, e.g. changes in environment parameters like friction or gravity \cite{OOD4,OOD5}, and are dealt with using forms of uncertainty estimation. Another form of OOD execution, one that is undetectable by uncertainty estimation, is a shift or perturbation in transition dynamics, referring to the definition of OOD provided in \cite{OOD2, ood101}, wherein the detectors aim to compare the distribution of transitions with a known transition function. However, a weakness of techniques that address the transition dynamics is that they are unable to account for epistemic uncertainty in their prediction \cite{OOD3}. Lastly, within the aforementioned context of offline-RL, OOD execution refers to a difference in the encountered state distribution from the training distribution \cite{offline1, uncer3}.

To summarize, we note that prior works on OOD detection within RL: \textbf{(1)} have incomplete definitions of OOD execution which lead to solutions that are unable to address general OOD execution in RL; \textbf{(2)} are unable to place guarantees on the outcome of OOD detection, which has consequences when designing safety-critical systems. We address these issues in our work by building on the definition of OOD described in \cite{OOD2}. The architecture of our detector utilizes CVAEs to learn the transition distribution, which we can use to predict the likelihood of a transition occurring and identify OOD execution. We apply a conformal prediction on the latent distribution, which allows us to guarantee OOD predictions.

\subsection{OOD Detection with VAEs}

Studies making use of VAEs for OOD detection exist outside of RL. These are grouped into: \textbf{(1)} studies using reconstruction error \cite{OODVAE1,OODVAE2,OODVAE11,OODVAE9}; \textbf{(2)} studies using the latent distributions to calculate likelihood \cite{OOD1,OODVAE12,OODVAE3}; \textbf{(3)} studies classifying the latent embeddings \cite{OODVAE5,OODVAE6}. A subset of these techniques provide confidence guarantees on OOD detection utilizing ICP \cite{OODVAE7,OODVAE8,OODVAE5}. However, adapting these techniques to RL pose a challenge as transition distributions are rarely Gaussian and ICP tends to be conservative in its predictions. This is another issue that we attempt to solve.


\section{Defining Out-Of-Distribution in MDPs} \label{defsec}
Potential reasons for OOD execution range from insufficient training to the agent being deployed in a significantly different environment. To develop our framework of OOD execution detection, we frame OOD execution in RL within the context of MDPs. Formally, the causes of OOD execution are grouped into two categories:

\begin{enumerate}
    \item \textbf{State-Based OOD Execution} describes a scenario where the learner generalizes a policy over an MDP with high uncertainty. The cause of this can be epistemic, wherein the learner is insufficiently trained over certain states. Similarly, the cause of this can be aleatoric, wherein additional experience does not reduce overall uncertainty. Using an MDP model, state-based aleatoric uncertainty is represented through high entropy transition functions where determining an action is unlikely to change the outcome, e.g. for some $s \in S, \; \; T(s_1 | s, a) = T(s_2 | s,a), \; \; \forall s_1, s_2 \in S, \;\; \forall a \in A$. The implication of state-based OOD execution is that there are some states over which the synthesized policy is uncertain and traversing them may result in OOD/unexpected outcomes. A practical solution suggested in \cite{uncer18,uncer16} is to avoid these states.
    
    \item \textbf{Environment-Based OOD Execution} describes a scenario where the learner, having generalized a policy within a training environment, is deployed in an unfamiliar environment. To distinguish this from state-based OOD execution, using an MDP model, we assume the state and action space, $S$ and $A$ respectively, are recognized as the same by the agent. However, the transition and reward functions, $T$ and $R$, may be different, resulting in unexpected outcomes from the execution of a policy that is held to be quite certain in the training environment. These could be the results of perturbations to input-output or systemic errors or changes to the environment/domain shifts \cite{OOD2, ood101}. Aleatoric uncertainty of this nature can be expected to some extent when deploying in real-life, however the objective of detecting this type of OOD execution is to halt learning-enabled systems from running if they demonstrate significantly different performances from training.
\end{enumerate}

\noindent To compare the types of OOD intuitively, the RL agent is aware of state-based OOD scenarios, but is unaware of environment-based OOD scenarios prior to deployment. This stems from the invariance assumption of learning, i.e. the training environment is representative of the deployment environment, which is contrary to the premise of environment-based OOD detection.

\begin{algorithm}[tb]
\caption{CVAE Training}
\label{algo1}
\textbf{Input}: Learnt policy, $\Pi : S \rightarrow A$; \\
\textbf{Parameter}:  MDP, $(S, A, T, R, \gamma)$; ensemble size, $N$; \\
\begin{algorithmic}[1] 
\vspace{-.5cm}
\STATE Initialize CVAE models $\{C_1 ... C_N\}$ with corresponding weights $\{ \theta_1 ... \theta_N 
\}$
\STATE Initialize variables $s_1, s_2$ to a state $s \in S$
\WHILE{\textit{Training}}
\STATE $s_2 \leftarrow s \sim T(s | s_1, \Pi(s_1))$ 
\STATE Compute CVAE loss for input, $(s_2 | s_2)$ over $\{ C_1 ... C_N \}$
\STATE Update weights $\{ \theta_1 ... \theta_N \}$
\STATE $s_1 \leftarrow s_2$
\ENDWHILE
\STATE \textbf{return} $\{C_1 ... C_N\}$

\end{algorithmic}
\end{algorithm}

\subsection{Mathematically Defining OOD Execution}
Within our problem scope, an assumption we make is that the state-wise reward, $\mathcal{R} : S \rightarrow \mathbb{R}$ is designed to be invariant between the training and deployment state-spaces. However, $R : S \times A \rightarrow \mathbb{R}$ is dependent on the transition function, $T$, and may not be the same between the two environments. Note that the policy learnt, $\Pi : S \rightarrow A$, is also dependent on $T$, though the learner might not have explicit knowledge of $T$. State-based OOD execution occurs when the learner executes an uncertain transition, and environment-based OOD execution occurs when there is a change in the transition function over part of the state-space. Therefore, constructing an OOD detector conditioned upon the transition function is the apparent solution. 

Let $T_t$ be the empirical transition function experienced by the learner during training, let $T_d$ be the transition function experienced during deployment and let $\delta$ be a confidence parameter describing the tolerance to OOD execution. A probabilistic OOD detection condition that compares the difference in distributions is provided in Inequality \eqref{eq4}. A practical application of this would consider the likelihood of the transition $s_1 \rightarrow s_2$, $s_1, s_2 \in S$, occurring under $\Pi(s_1)$, and consider it an example of OOD execution if the probability is low.

\begin{align} \label{eq4}
    \mathbb{P} \left(  T_t(s_2 | s_1, \Pi(s_1)) \neq T_d(s_2 | s_1, \Pi(s_1))  \right) > \delta
\end{align}

\noindent However, we note that an attempt to directly compare $T_t$ with $T_d$ is insufficient. While Inequality \eqref{eq4} is able to recognize shifts in the transition function as well as insufficient experience, state-based aleatoric uncertainty would go unrecognized. High-entropy transition functions would still be recognized as ID even with the presence of high uncertainty. We modify this in Inequality \eqref{eq3}, where we incorporate the detection of state-based aleatoric uncertainty into our detection condition. Assuming that the detection algorithm is being executed following a transition from $s_1$ under $\Pi(s_1)$ on $T_d$, its objective is to assess the probability that the sample drawn from $T_d$, i.e. $s_2 \sim T_d(\cdot | s_1, \Pi(s_1))$, is unequal to the expected value of $T_t$. If the value exceeds a threshold $\delta$, it is an instance of OOD execution.

\begin{algorithm}[tb]
\caption{ICP Calibration}
\label{algo2}
\textbf{Input}: Learnt policy, $\Pi: S \rightarrow A$; trained CVAE ensemble, $\mathcal{C}_N$; \\
\textbf{Parameter}:  MDP, $(S, A, T, R, \gamma)$; calibration size, $M$; \\
\begin{algorithmic}[1] 
\vspace{-0.5cm}
\STATE Initialize variable $m \leftarrow 0$
\STATE Initialize set $RECON \leftarrow \{ \}$
\STATE Initialize variables $s_1, s_2$ to a state $s \in S$
\WHILE{$m < M$}
\STATE $s_2 \leftarrow s \sim T(s | s_1, \Pi(s_1))$ 
\STATE Compute reconstruction errors $\{r^*_1 ... r^*_N \}$ for $\mathcal{C}_N(s_2 | s_1)$
\STATE Join $\{r^*_1 ... r^*_N \}$ to $RECON$
\STATE $s_1 \leftarrow s_2$ 
\STATE $m \leftarrow m+1$
\ENDWHILE
\STATE Sort $RECON$
\STATE \textbf{return} $RECON$

\end{algorithmic}
\end{algorithm}

\begin{algorithm}[tb]
\caption{OOD Detection}
\label{algo3}
\textbf{Input}: Transition pair, $s_1, s_2 \in S$; trained CVAE ensemble, $\mathcal{C}_N$; sorted (ascending) calibration set, $\{ r_1 ... r_{M \times N} \}$;\\
\textbf{Parameter}:  Confidence, $\delta \in (0, 1)$; \\
\begin{algorithmic}[1] 
\vspace{-0.5cm}
\STATE Initialize set $\Gamma \leftarrow \{ \}$
\STATE Compute reconstruction errors $\{r^*_1 ... r^*_N \}$ for $\mathcal{C}_N(s_2 | s_1)$
\FOR{$r^*_i \in \{r^*_1 ... r^*_N \}$}
\IF{$r^*_i < r_{\lfloor \delta \times M \times N \rfloor}$}
\STATE Append $r^*_i$ to $\Gamma$
\ENDIF
\ENDFOR

\IF{$\Gamma = \phi $}
\STATE \textbf{return} $TRUE$
\ELSE
\STATE \textbf{return} $FALSE$
\ENDIF

\end{algorithmic}
\end{algorithm}

\begin{align} \label{eq3}
    \mathbb{P} \left( \underset{T_t(s | s_1, \Pi(s_1))}{\mathbb{E}} [s] \neq s_2  \right) > \delta
\end{align} 

\begin{table}[t]
\centering
\small
\begin{tabular}{|c|c|c|c|}
\hline
\multirow{2}{.3em}{\#} & Uncertainty & Empirical Transition & Ground-\\
     &  Estimation & Probability &  Truth\\
    \hline
    \textbf{A} & High Var. / OOD & Low Prob. / OOD & OOD \\
    \hline
    \textbf{B} & High Var. / OOD & High Prob. / ID & ID\\
    \hline
    \textbf{C} & Low Var. / ID & Low Prob. / OOD & OOD \\
    \hline
    \textbf{D} & Low Var. / ID & High Prob. / ID & ID \\
    \hline
\end{tabular}
\caption{Comparing different OOD detection methods with ground truth}
\label{table1}
\end{table}

\noindent The condition described in Inequality \eqref{eq3}, which we call \emph{empirical transition likelihood}, encapsulates both state-based and environment-based OOD execution. Thus, we compare the difference between the expected empirical transitions and the sampled states from the deployment transition distribution. In practice, for nominal discrete states, the expected value of $T_t$ can be substituted by the most frequent transition.

\subsection{Contrasting Empirical Transition Likelihood with Prior Definitions}

While similar to the approach of learning transition dynamics in PEDM \cite{OOD2} and DEXTER \cite{ood101}, our OOD detection condition accounts for state-based as well as environment-based OOD execution by making a distinction between the distribution of states encountered by the agent during training and the actual distribution of states. Noting that state-based OOD execution is distinct from the perturbation-based OOD execution defined in the aforementioned studies, and more akin to learning uncertainty, we also note that existing transition-based methods are unable to detect this.

We also briefly contrast this with the uncertainty estimation methods presented in \cite{uncer12}, which determine states with high variance in value, $V(s)$, as OOD. Assume that a threshold exists for value variance to determine \textit{low variance} and \textit{high variance} states, and a similar threshold exists in empirical transition likelihood for \textit{low likelihood} and \textit{high likelihood} states. Also assuming there exist two OOD detectors, the first making predictions using value uncertainty and the second making predictions using empirical transition likelihood, the outcome of OOD detection of any transition that takes place will be one of either \textbf{A}, \textbf{B}, \textbf{C} or \textbf{D}, as depicted in Table \ref{table1}.

Noting the outcomes of each scenario in Table \ref{table1}, scenarios \textbf{A} and \textbf{D} being OOD (state-based epistemic uncertainty) and ID, respectively, is self-explanatory. However, when the predictions made by both detectors are contradictory, i.e. \textbf{B} and \textbf{C}, assigning an outcome requires rationalizing. If a transition has a low likelihood of occurring but the value estimator is confident in its prediction, as in the case of \textbf{C}, two scenarios are possible: \textbf{(1)} state-based aleatoric uncertainty is present, wherein the transition distribution has a high entropy; \textbf{(2)} it is a case of environment-based OOD execution that is not detectable by the uncertainty estimator, which has been trained to estimate value variance within the training environment. 

Scenario \textbf{B} occurs when the transition is ID, however there is significant uncertainty in the value. Noting the value definition in Equation \eqref{valuedef}, the only feasible way for this to occur is if the subsequent transitions place the learner in scenario \textbf{A}, with the high variance in the value of a future transition propagating back to the current scenario, \textbf{B}. Thus, while the current transition is not an instance of OOD execution, it may lead to OOD execution at a future timestep.


\section{Detecting OOD with High-Confidence}

\noindent The aim of this study is to detect OOD execution with high confidence using the detection condition in Inequality \eqref{eq3}. To do so, we aim to learn the empirical transition dynamics using CVAEs. Prior works have aimed to learn transition dynamics through probabilistic dynamics models \cite{OOD2}. The use of CVAEs presents two significant advantages. Firstly, the architecture is agnostic to the RL algorithm used, and decoupling learning from OOD detection is a practical advantage that allows for modification of learning-enabled components as required by the system. Secondly, autoencoder models allow for the use of reconstruction error as an alternative to likelihood, where reconstruction accuracy is dependent on the frequency with which an instance is observed. This incorporates state-based epistemic uncertainty into the detection, which probabilistic dynamics models do not recognize.

\begin{figure}[t]
\centering
\includegraphics[width=0.45\textwidth]{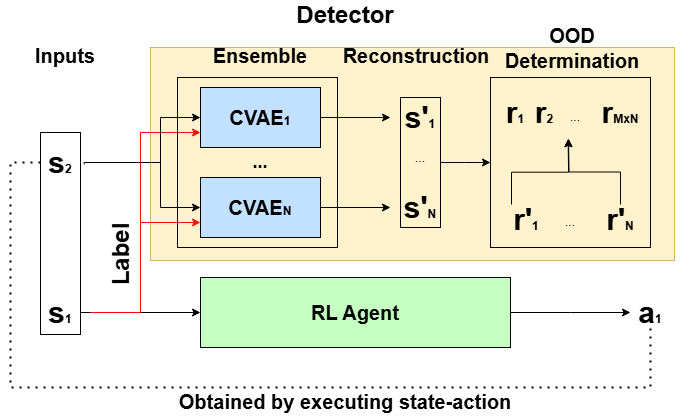}
\caption{Depicted is the OOD detection architecture, consisting of (green) a trained RL agent and (yellow) the detector consisting of an ensemble of $N$ CVAEs (blue). Denoting the calibration set using $\{ r_1 ... r_{M \times N} \}$, the CVAE ensemble produces $N$ reconstructions of state $s_2$, $\{ s'_1 ... s'_N\}$, conditioned on $s_1$. The reconstruction error with each of $s'_1 ... s'_N$ is denoted using $r'_1 ... r'_N$ and is compared with the elements of the calibration set.}
\label{diagram}
\end{figure}

In order to learn the empirical transition likelihood given a policy, $\Pi(s)$, we train a CVAE ensemble to approximate the distribution of $s_2 \in S$, $T(s_2 | s_1, \Pi(s_1))$, given state $s_1 \in S$. That is, upon executing a transition, each CVAE is trained to encode and decode $s_2$ given the label $s_1$. As such, denoting the latent variable using $z$, the input using $s_2$ and the label using $s_1$, the objective of each CVAE is to learn $\mathcal{P}(z | s_2, s_1)$. The implication of using CVAEs is that the latent distribution learnt for each label is trained to approximate the empirical transition distribution using a Gaussian model. Furthermore, samples drawn from the latent distribution with high probability are more likely to be decoded as states that are ID. As such, building an OOD detector using the positional encoding of $z$, corresponding with pair $(s_1, s_2)$, is a viable method as $\mathcal{P}(z)$ is a known Gaussian distribution. However, there is no guarantee on how well $\mathcal{Q}_\theta(z | s_2, s_1)$ approximates $\mathcal{P}(z | s_2, s_1)$. We depict the architecture of our OOD detection system in Figure \ref{diagram}.

An observed state, $s_1$, is input to RL model, which acts and transitions to $s_2$. The values $s_1$ and $s_2$ are input to the CVAE models, with $s_1$ as the label, and the mean squared errors of the reconstructions of $s_2$ are measured. We detail the training method (agnostic of the RL algorithm used) in Algorithm \ref{algo1}, where we train an ensemble consisting of $N$ CVAEs initialized with random weights, to produce reconstructions of $s_2$. The OOD detector is constructed using ICP, described in Algorithm \ref{algo2}. The reconstruction errors experienced during $M$ steps within the training environment, from each of $N$ CVAEs in the ensemble, is used to build a calibration set of size $M \times N$. 

Algorithm \ref{algo3} describes the OOD detection process during deployment. In order to measure the ID-ness of the reconstruction errors by the CVAE ensemble, we compare them against reconstruction errors observed during the calibration process for the detector. We use $\delta$ to pick the threshold value from the calibration set, $r_{ \lfloor \delta \times M \times N \rfloor}$, which guarantees a true positive OOD detection rate of $1-\delta$. Noting that there are $N$ CVAEs, each detection produces a set of $N$ reconstruction errors. As ICP produces a set prediction, we can treat the non-conformity of each score in the reconstruction set as a $\phi$ prediction, i.e. OOD. 

The use of an ensemble in this architecture provides two benefits. Firstly, each CVAE is able to learn a different approximation of $\mathcal{P}(s_2 | s_1)$, allowing the ensemble to, overall, better approximate it. Secondly, to guarantee a high true positive OOD detection rate, the false positive rate is also high. The use of multiple CVAEs reduces the probability of a false positive detection given a guaranteed true positive rate as there are more `classes' for the set predictor to not conform with.


\section{Experiments} 


The experiments run in this study are to evaluate the performance of the OOD detector. We utilize the \textbf{LunarLander}, \textbf{Ant}, \textbf{CartPole}, and \textbf{Pendulum} environments from OpenAI Gym \cite{gym}. We adapt benchmarks proposed in prior works \cite{uncer1,bench1,OOD5,OOD2}. Our experiment consists of the following steps:

\begin{enumerate}
    \item Train a Deep RL agent and OOD detection system in a continuous MDP setting with parameters $(S, A, T, R, \gamma)$ and confidence threshold $\delta$.
    \item The agent and detection system are deployed in a continuous MDP setting $(S, A, T^*, R, \gamma)$, where the transition function has been altered.
    \item Upon detecting OOD execution an episode is prematurely ended and true/false positive OOD detection is registered.
    \item The previous steps are repeated using different environments and alterations to the transition function.
\end{enumerate}

\noindent We train our OOD detection system using an Advantage Actor-Critic model \cite{a2c}. The metric we use to assess performance is the area under the receiver operating curve (AUC). When altering the transition function for each episode of the experiment, based on prior work, we make one of two alterations: \textbf{(1)} we `corrupt' the environment parameters, e.g. in the Pendulum environment set a different gravity when deployed than what was trained on \cite{uncer1}; \textbf{(2)} we add a random perturbation to the action \cite{OOD2}. Table \ref{table2} displays the full list of potential parameter changes.

We aim to detect two types of OOD executions, \textit{state-based} and \textit{environment-based}. However, the alterations proposed in Table \ref{table2} only result in environment-based OOD execution. Therefore, we also deploy in environments that are identical to the training environment, in order to test for state-based OOD execution as well as false positives. Testing for state-based OOD execution requires knowledge of the training distribution as well as the transition function. Therefore, we assess performance in the detection of state-based OOD execution by using known certain states and known uncertain states, i.e. frequently visited states with low-entropy transition distributions wrt. actions vs. rarely seen states with high-entropy transition distributions wrt. actions, and the detection results over transitions to/from these states. For example, high velocity states in the Pendulum and LunarLander environments are rarely encountered on the trained policy and known to be OOD.

\begin{table}[t]
\centering
\scriptsize
\begin{tabular}{|c|c|c|c|}
\hline
    \multirow{2}{5.5em}{\textbf{Environment}} & \multirow{2}{5.5em}{\textbf{Parameters}} & \textbf{Training} & \textbf{Alteration}\\
      \; &  \;   & \textbf{Value}    & \textbf{Range} \\
    \hline
    \multirow{3}{5.5em}{\textbf{LunarLander}} & Gravity    & -10.0 & [-0.1, -12.0] \\
                         & WindPower  & 0.3   & [0.0, 40.0]   \\ 
                         & Turbulence & 0.3   & [0.0, 40.0]   \\
    \hline
    \multirow{1}{5.5em}{\textbf{Ant}} & Torque Scaling    & 1.0 & [0.5, 1.5] \\
    \hline
     \multirow{4}{5.5em}{\textbf{Pendulum}}    & Gravity    & 10.0  & [0.1, 50.0]   \\
                         & Mass       & 1.0   & [0.1, 10.0]   \\
                         & Length     & 1.0   & [0.1, 5.0]    \\
                         & Max. Velocity  & 8.0   & [1.0, 20.0]   \\
    \hline
    \multirow{4}{5.5em}{\textbf{CartPole}}    & Gravity    & 10.0  & [0.1, 50.0]   \\
                         & CartMass   & 1.0   & [0.1, 10.0]   \\
                         & PoleLength & 0.5   & [0.1, 5.0]    \\
                         & PoleMass   & 0.1   & [0.01, 1.0]   \\
    \hline
\end{tabular}
\caption{Possible Alterations to Environment}
\label{table2}
\end{table}

\subsection{Results and Analysis}

\begin{table*}[h!]
\centering
\scriptsize
\begin{tabular}{|c|c|c|c|c|c|c|}
\hline
    \textbf{Environment} & \textbf{Model} & \textbf{Overall AUC} & \textbf{State AUC} & \textbf{Env. AUC} & \textbf{Env. (1) AUC} & \textbf{Env. (2) AUC}\\
    \hline
    \multirow{5}{5.5em}{\textbf{LunarLander}} & COTD (Ours)    & \textbf{0.807} & \textbf{0.852} & \textbf{0.762} & \textbf{0.714} & \textbf{0.809}\\
                                              & PEDM            & 0.648 & - & 0.648 & 0.632 & 0.663\\ 
                                              & DEXTER            & 0.657 & - & 0.657 & 0.581 & 0.733\\ 
                                              & MC-DC          & 0.589 & 0.610 & - & 0.548 & - \\
                                              & UBOOD         & 0.761 & 0.850 & - & 0.583 & - \\
    \hline
    \multirow{5}{5.5em}{\textbf{Ant}} & COTD (Ours)           & \textbf{0.875} & \textbf{0.936} & \textbf{0.814} & \textbf{0.860} & 0.768 \\
                                              & PEDM          & 0.799 & -     & 0.799 & 0.853 & 0.744 \\ 
                                              & DEXTER        & 0.671 & -     & 0.671 & 0.512 & \textbf{0.829} \\ 
                                              & MC-DC         & 0.624 & 0.606 & -     & 0.641 & - \\
                                              & UBOOD         & 0.627 & 0.877 & -     & 0.376 & - \\
    \hline
    \multirow{5}{5.5em}{\textbf{Pendulum}}    & COTD (Ours)    & \textbf{0.765} & 0.797 & \textbf{0.733} & 0.696 & 0.770\\
                                              & PEDM            & 0.706 & -  & 0.706 & \textbf{0.700} & 0.712\\ 
                                              & DEXTER            & 0.697 & - & 0.697 & 0.579 & \textbf{0.814}\\ 
                                              & MC-DC          & 0.607 & 0.598 & - & 0.625 & - \\
                                              & UBOOD         & 0.753 & \textbf{0.823} & - & 0.614 & - \\
    \hline
    \multirow{5}{5.5em}{\textbf{CartPole}}    & COTD (Ours)    & \textbf{0.759} & \textbf{0.771} & \textbf{0.747} & 0.737 & 0.757\\
                                              & PEDM            & 0.691 & - & 0.691 & 0.647 & 0.733\\ 
                                              & DEXTER            & 0.706 & - & 0.706 & 0.633 & \textbf{0.779}\\ 
                                              & MC-DC          & 0.713 & 0.694 & - & \textbf{0.750} & - \\
                                              & UBOOD         & 0.725 & 0.755 & - & 0.669 & - \\
    \hline
\end{tabular}
\caption{Experiment Results}
\label{table3}
\end{table*}

To obtain our results, for each of the LunarLander, Ant, Pendulum and CartPole environments, we tested our OOD detector in a total of 2000 episodes; 1000 episodes in which the deployment environments and transition functions were identical to that of  training, 500 episodes in which the actions were randomly perturbed and 500 episodes in which the environment parameters were set to random values as described in Table \ref{table2}. These results are obtained by setting the CVAE ensemble size to 5 and are described in Table \ref{table3}, where we compare our detector, CVAE-based OOD Transition Detector (COTD), against existing the existing state-of-art OOD detectors for RL: Detection Via Extraction of Time Series Representation (DEXTER) \cite{ood101}, Probabilistic Ensemble Dynamics Model based Detectors (PEDM) \cite{OOD2}, Monte Carlo DropConnect based Detectors (MC-DC) \cite{OOD5} and Uncertainty-Based OOD Classification with Monte Carlo Dropout (UBOOD) \cite{uncer1}.

The \textbf{State AUC} column is the evaluation of the models in episodes where the deployment environment is identical to the training environment and the objective is to detect state-based OOD execution. The \textbf{Env. AUC}, \textbf{Env. (1) AUC} and \textbf{Env. (2) AUC} columns summarize the detection performance on environment-based OOD execution, where \textbf{Env. (1) AUC} considers performance against parameter changes, \textbf{Env. (2) AUC} considers performance against transition perturbations and \textbf{Env. AUC} is a combined metric that averages both of the aforementioned wherever they both exist. When considering the false positive rate for this metric, we utilize the false positive detections on the unaltered environments. Note that some entries in Table \ref{table3} are empty. This is because the respective models are not intended to be used to detect that particular kind of OOD execution, e.g. UBOOD is unable to detect OOD execution due to random perturbations, and thus, the cell \textbf{Env. (2) AUC} is empty for UBOOD. As such, comparing models for their performance in non-intended purposes is inappropriate.

From Table \ref{table3}, we can claim that the proposed COTD outperforms the other models in most of the experiments. Specifically, for \textbf{Env. AUC}, COTD outperforms other methods by at least  $17\%$, $4\%$, and $8\%$ in the LunarLander, Pendulum, and CartPole settings, respectively. Also, for \textbf{Overall AUC}, the COTD outperforms other methods by at least  $6\%$, $7\%$ $2\%$, and $4\%$ in the LunarLander, Ant, Pendulum, and CartPole settings, respectively. A point to note is that these performance gains in OOD detection exist across multiple types of OOD execution that we test our agent within and that detection is conducted with a confidence guarantee, i.e. while guaranteeing the true positive detection rate using ICP. We note that the types of transition distribution learnt by our algorithm is limited to a Gaussian distribution. However, by using multiple Gaussians in the form of a CVAE ensemble, we are able to better approximate the transition distribution while limiting the impact of complexity on our detector. This is reflected by the consistent performance across environments as well as OOD execution type.

To conclude our analysis, we comment on the choice of COTD hyperparameters. The ensemble size of 5 was made by optimizing over the false positive rate as the guarantee over true positive rate exists due to the choice of $\delta$. We posit that increasing the ensemble size reduces the false positive rate as ICP allows the data instance more classes to conform with when there are more CVAEs, though additional increases to the ensemble size result in reduced gains in the false positive rate as well as decreases in detection speed. In addition, letting $s_1$ be an initial state, $s_2$ be a state transitioned from $s_1$ and $s_3$ be a state transition from $s_2$, training the CVAE using the pair $(s_3 | s_1)$, instead of $(s_2 | s_1)$ as in this study, results in more accurate detections at the cost of timesteps taken to detect OOD execution. However, drawing out the detection steps may be necessary in continuous-time MDPs where state transitions are imperceptible or cases where transition distributions are non-approximable.

\section{Conclusions and Further Work}
The main contributions of this study are in providing a comprehensive definition of OOD execution in RL as well as designing an OOD detection algorithm with probabilistic guarantees on detection. We identify and distinguish between the types of OOD execution, i.e. state-based vs. environment-based, utilizing the characteristics of each to formulate a general definition for OOD execution in RL within Section \ref{defsec}. Based on this, we construct a detector that aims to model the transition kernel for each state and use the reconstruction error of the CVAE to identify OOD execution. For a user defined parameter, $\delta \in [0, 1]$, the confidence with which a detection is conducted is specified.

Subsequently, We demonstrate that our model, COTD, outperforms prior works using experimental results. The implication of this is that transition distribution approximation can be more robust than uncertainty estimation methods when detecting OOD execution within RL. Similarly, our study also extends the results of prior works on OOD execution detection using knowledge of the transition function to allow for the detection of state-based OOD execution. We intend for further work to investigate the tradeoff between accuracy and detection speed through hyperparameter tuning, i.e. ensemble size and detection steps, as well as improving on the false positive detection rate with a probabilistic guarantee on the true positive rate.

\section{Acknowledgements}
This research is supported by the National Research Foundation, Singapore and DSO National Laboratories under the AI Singapore Programme (AISG Award No: AISG2-RP-2020-017). This research is also supported by the National Research Foundation, Prime Minister’s Office, Singapore under its Campus for Research Excellence and Technological Enterprise (CREATE) programme through the programme DesCartes. This research was also supported by MoE, Singapore, through the Tier-2 grant MOE2019-T2-2-040. 

\bibliography{aaai25}

\begin{thebibliography}{43}
\providecommand{\natexlab}[1]{#1}

\bibitem[{An et~al.(2021)An, Moon, Kim, and Song}]{uncer14}
An, G.; Moon, S.; Kim, J.-H.; and Song, H.~O. 2021.
\newblock Uncertainty-based offline reinforcement learning with diversified q-ensemble.
\newblock \emph{Advances in neural information processing systems}, 34: 7436--7447.

\bibitem[{An and Cho(2015)}]{OODVAE1}
An, J.; and Cho, S. 2015.
\newblock Variational autoencoder based anomaly detection using reconstruction probability.
\newblock \emph{Special lecture on IE}, 2(1): 1--18.

\bibitem[{Bai et~al.(2022)Bai, Wang, Yang, Deng, Garg, Liu, and Wang}]{offline2}
Bai, C.; Wang, L.; Yang, Z.; Deng, Z.; Garg, A.; Liu, P.; and Wang, Z. 2022.
\newblock Pessimistic bootstrapping for uncertainty-driven offline reinforcement learning.
\newblock \emph{arXiv preprint arXiv:2202.11566}.

\bibitem[{Brockman et~al.(2016)Brockman, Cheung, Pettersson, Schneider, Schulman, Tang, and Zaremba}]{gym}
Brockman, G.; Cheung, V.; Pettersson, L.; Schneider, J.; Schulman, J.; Tang, J.; and Zaremba, W. 2016.
\newblock Openai gym.
\newblock \emph{arXiv preprint arXiv:1606.01540}.

\bibitem[{Brunke et~al.(2022)Brunke, Greeff, Hall, Yuan, Zhou, Panerati, and Schoellig}]{uncer15}
Brunke, L.; Greeff, M.; Hall, A.~W.; Yuan, Z.; Zhou, S.; Panerati, J.; and Schoellig, A.~P. 2022.
\newblock Safe learning in robotics: From learning-based control to safe reinforcement learning.
\newblock \emph{Annual Review of Control, Robotics, and Autonomous Systems}, 5: 411--444.

\bibitem[{Cai and Koutsoukos(2020)}]{OODVAE7}
Cai, F.; and Koutsoukos, X. 2020.
\newblock Real-time out-of-distribution detection in learning-enabled cyber-physical systems.
\newblock In \emph{2020 ACM/IEEE 11th International Conference on Cyber-Physical Systems (ICCPS)}, 174--183. IEEE.

\bibitem[{Clements et~al.(2019)Clements, Van~Delft, Robaglia, Slaoui, and Toth}]{uncer12}
Clements, W.~R.; Van~Delft, B.; Robaglia, B.-M.; Slaoui, R.~B.; and Toth, S. 2019.
\newblock Estimating risk and uncertainty in deep reinforcement learning.
\newblock \emph{arXiv preprint arXiv:1905.09638}.

\bibitem[{Da~Silva et~al.(2020)Da~Silva, Hernandez-Leal, Kartal, and Taylor}]{uncer18}
Da~Silva, F.~L.; Hernandez-Leal, P.; Kartal, B.; and Taylor, M.~E. 2020.
\newblock Uncertainty-aware action advising for deep reinforcement learning agents.
\newblock In \emph{Proceedings of the AAAI conference on artificial intelligence}, volume~34, 5792--5799.

\bibitem[{Danesh and Fern(2021)}]{bench1}
Danesh, M.~H.; and Fern, A. 2021.
\newblock Out-of-Distribution Dynamics Detection: RL-Relevant Benchmarks and Results.
\newblock \emph{arXiv preprint arXiv:2107.04982}.

\bibitem[{Denouden et~al.(2018)Denouden, Salay, Czarnecki, Abdelzad, Phan, and Vernekar}]{OODVAE2}
Denouden, T.; Salay, R.; Czarnecki, K.; Abdelzad, V.; Phan, B.; and Vernekar, S. 2018.
\newblock Improving reconstruction autoencoder out-of-distribution detection with mahalanobis distance.
\newblock \emph{arXiv preprint arXiv:1812.02765}.

\bibitem[{Feinberg and Shwartz(2012)}]{markov}
Feinberg, E.~A.; and Shwartz, A. 2012.
\newblock \emph{Handbook of Markov decision processes: methods and applications}, volume~40.
\newblock Springer Science \& Business Media.

\bibitem[{Gardille and Ahmad(2023)}]{OOD3}
Gardille, A.; and Ahmad, O. 2023.
\newblock Towards Safe Reinforcement Learning via OOD Dynamics Detection in Autonomous Driving System (Student Abstract).
\newblock In \emph{Proceedings of the AAAI Conference on Artificial Intelligence}, volume~37, 16216--16217.

\bibitem[{Haider et~al.(2024)Haider, Roscher, Herd, Schmoeller~Roza, and Burton}]{haider10}
Haider, T.; Roscher, K.; Herd, B.; Schmoeller~Roza, F.; and Burton, S. 2024.
\newblock Can you trust your Agent? The Effect of Out-of-Distribution Detection on the Safety of Reinforcement Learning Systems.
\newblock In \emph{Proceedings of the 39th ACM/SIGAPP Symposium on Applied Computing}, 1569--1578.

\bibitem[{Haider et~al.(2023)Haider, Roscher, Schmoeller~da Roza, and G{\"u}nnemann}]{OOD2}
Haider, T.; Roscher, K.; Schmoeller~da Roza, F.; and G{\"u}nnemann, S. 2023.
\newblock Out-of-Distribution Detection for Reinforcement Learning Agents with Probabilistic Dynamics Models.
\newblock In \emph{Proceedings of the 2023 International Conference on Autonomous Agents and Multiagent Systems}, 851--859.

\bibitem[{Haider et~al.(2021)Haider, Roza, Eilers, Roscher, and G{\"u}nnemann}]{OOD4}
Haider, T.; Roza, F.~S.; Eilers, D.; Roscher, K.; and G{\"u}nnemann, S. 2021.
\newblock Domain Shifts in Reinforcement Learning: Identifying Disturbances in Environments.
\newblock In \emph{AISafety@ IJCAI}.

\bibitem[{Kahn et~al.(2017)Kahn, Villaflor, Pong, Abbeel, and Levine}]{uncer11}
Kahn, G.; Villaflor, A.; Pong, V.; Abbeel, P.; and Levine, S. 2017.
\newblock Uncertainty-aware reinforcement learning for collision avoidance.
\newblock \emph{arXiv preprint arXiv:1702.01182}.

\bibitem[{Kingma and Welling(2013)}]{VAE}
Kingma, D.~P.; and Welling, M. 2013.
\newblock Auto-encoding variational bayes.
\newblock \emph{arXiv preprint arXiv:1312.6114}.

\bibitem[{Kumar et~al.(2020)Kumar, Zhou, Tucker, and Levine}]{offline1}
Kumar, A.; Zhou, A.; Tucker, G.; and Levine, S. 2020.
\newblock Conservative q-learning for offline reinforcement learning.
\newblock \emph{Advances in Neural Information Processing Systems}, 33: 1179--1191.

\bibitem[{Li et~al.(2021)Li, Gupta, Reddy, Pong, Zhou, Yu, and Levine}]{uncer16}
Li, K.; Gupta, A.; Reddy, A.; Pong, V.~H.; Zhou, A.; Yu, J.; and Levine, S. 2021.
\newblock Mural: Meta-learning uncertainty-aware rewards for outcome-driven reinforcement learning.
\newblock In \emph{International conference on machine learning}, 6346--6356. PMLR.

\bibitem[{Lockwood and Si(2022)}]{uncer9}
Lockwood, O.; and Si, M. 2022.
\newblock A review of uncertainty for deep reinforcement learning.
\newblock In \emph{Proceedings of the AAAI Conference on Artificial Intelligence and Interactive Digital Entertainment}, volume~18, 155--162.

\bibitem[{L{\"u}tjens, Everett, and How(2019)}]{uncer4}
L{\"u}tjens, B.; Everett, M.; and How, J.~P. 2019.
\newblock Safe reinforcement learning with model uncertainty estimates.
\newblock In \emph{2019 International Conference on Robotics and Automation (ICRA)}, 8662--8668. IEEE.

\bibitem[{Ma, Jayaraman, and Bastani(2021)}]{offline3}
Ma, Y.; Jayaraman, D.; and Bastani, O. 2021.
\newblock Conservative offline distributional reinforcement learning.
\newblock \emph{Advances in neural information processing systems}, 34: 19235--19247.

\bibitem[{Mai, Mani, and Paull(2022)}]{uncer5}
Mai, V.; Mani, K.; and Paull, L. 2022.
\newblock Sample efficient deep reinforcement learning via uncertainty estimation.
\newblock \emph{arXiv preprint arXiv:2201.01666}.

\bibitem[{Mnih et~al.(2016)Mnih, Badia, Mirza, Graves, Lillicrap, Harley, Silver, and Kavukcuoglu}]{a2c}
Mnih, V.; Badia, A.~P.; Mirza, M.; Graves, A.; Lillicrap, T.; Harley, T.; Silver, D.; and Kavukcuoglu, K. 2016.
\newblock Asynchronous methods for deep reinforcement learning.
\newblock In \emph{International conference on machine learning}, 1928--1937. PMLR.

\bibitem[{Mohammed and Valdenegro-Toro(2021)}]{OOD5}
Mohammed, A.~P.; and Valdenegro-Toro, M. 2021.
\newblock Benchmark for out-of-distribution detection in deep reinforcement learning.
\newblock \emph{arXiv preprint arXiv:2112.02694}.

\bibitem[{Nasvytis et~al.(2024)Nasvytis, Sandbrink, Foerster, Franzmeyer, and de~Witt}]{ood101}
Nasvytis, L.; Sandbrink, K.; Foerster, J.; Franzmeyer, T.; and de~Witt, C.~S. 2024.
\newblock Rethinking Out-of-Distribution Detection for Reinforcement Learning: Advancing Methods for Evaluation and Detection.
\newblock \emph{arXiv preprint arXiv:2404.07099}.

\bibitem[{Okada and Taniguchi(2020)}]{uncer8}
Okada, M.; and Taniguchi, T. 2020.
\newblock Variational inference mpc for bayesian model-based reinforcement learning.
\newblock In \emph{Conference on robot learning}, 258--272. PMLR.

\bibitem[{Prashant and Easwaran(2022)}]{OODVAE8}
Prashant, M.; and Easwaran, A. 2022.
\newblock PAC-Based Formal Verification for Out-of-Distribution Data Detection.
\newblock In \emph{2022 6th International Conference on System Reliability and Safety (ICSRS)}, 300--309. IEEE.

\bibitem[{Rahiminasab, Yuhas, and Easwaran(2022)}]{OODVAE6}
Rahiminasab, Z.; Yuhas, M.; and Easwaran, A. 2022.
\newblock Out of Distribution Reasoning by Weakly-Supervised Disentangled Logic Variational Autoencoder.
\newblock In \emph{2022 6th International Conference on System Reliability and Safety (ICSRS)}, 169--178. IEEE.

\bibitem[{Ramakrishna et~al.(2022)Ramakrishna, Rahiminasab, Karsai, Easwaran, and Dubey}]{OODVAE5}
Ramakrishna, S.; Rahiminasab, Z.; Karsai, G.; Easwaran, A.; and Dubey, A. 2022.
\newblock Efficient out-of-distribution detection using latent space of $\beta$-vae for cyber-physical systems.
\newblock \emph{ACM Transactions on Cyber-Physical Systems (TCPS)}, 6(2): 1--34.

\bibitem[{Ren et~al.(2019)Ren, Liu, Fertig, Snoek, Poplin, Depristo, Dillon, and Lakshminarayanan}]{OOD1}
Ren, J.; Liu, P.~J.; Fertig, E.; Snoek, J.; Poplin, R.; Depristo, M.; Dillon, J.; and Lakshminarayanan, B. 2019.
\newblock Likelihood ratios for out-of-distribution detection.
\newblock \emph{Advances in neural information processing systems}, 32.

\bibitem[{Sedlmeier et~al.(2019)Sedlmeier, Gabor, Phan, Belzner, and Linnhoff-Popien}]{uncer1}
Sedlmeier, A.; Gabor, T.; Phan, T.; Belzner, L.; and Linnhoff-Popien, C. 2019.
\newblock Uncertainty-based out-of-distribution classification in deep reinforcement learning.
\newblock \emph{arXiv preprint arXiv:2001.00496}.

\bibitem[{Sedlmeier et~al.(2020)Sedlmeier, M{\"u}ller, Illium, and Linnhoff-Popien}]{uncer7}
Sedlmeier, A.; M{\"u}ller, R.; Illium, S.; and Linnhoff-Popien, C. 2020.
\newblock Policy entropy for out-of-distribution classification.
\newblock In \emph{Artificial Neural Networks and Machine Learning--ICANN 2020: 29th International Conference on Artificial Neural Networks, Bratislava, Slovakia, September 15--18, 2020, Proceedings, Part II 29}, 420--431. Springer.

\bibitem[{Serr{\`a} et~al.(2019)Serr{\`a}, {\'A}lvarez, G{\'o}mez, Slizovskaia, N{\'u}{\~n}ez, and Luque}]{OODVAE12}
Serr{\`a}, J.; {\'A}lvarez, D.; G{\'o}mez, V.; Slizovskaia, O.; N{\'u}{\~n}ez, J.~F.; and Luque, J. 2019.
\newblock Input complexity and out-of-distribution detection with likelihood-based generative models.
\newblock \emph{arXiv preprint arXiv:1909.11480}.

\bibitem[{Sohn, Lee, and Yan(2015)}]{CVAE}
Sohn, K.; Lee, H.; and Yan, X. 2015.
\newblock Learning Structured Output Representation using Deep Conditional Generative Models.
\newblock In Cortes, C.; Lawrence, N.; Lee, D.; Sugiyama, M.; and Garnett, R., eds., \emph{Advances in Neural Information Processing Systems}, volume~28. Curran Associates, Inc.

\bibitem[{Vovk(2012)}]{ICP}
Vovk, V. 2012.
\newblock Conditional validity of inductive conformal predictors.
\newblock In \emph{Asian conference on machine learning}, 475--490. PMLR.

\bibitem[{Wang and Zou(2021)}]{uncer13}
Wang, Y.; and Zou, S. 2021.
\newblock Online robust reinforcement learning with model uncertainty.
\newblock \emph{Advances in Neural Information Processing Systems}, 34: 7193--7206.

\bibitem[{Wu, Huang, and Lv(2022)}]{uncer17}
Wu, J.; Huang, Z.; and Lv, C. 2022.
\newblock Uncertainty-aware model-based reinforcement learning: Methodology and application in autonomous driving.
\newblock \emph{IEEE Transactions on Intelligent Vehicles}, 8(1): 194--203.

\bibitem[{Wu et~al.(2021)Wu, Zhai, Srivastava, Susskind, Zhang, Salakhutdinov, and Goh}]{uncer6}
Wu, Y.; Zhai, S.; Srivastava, N.; Susskind, J.; Zhang, J.; Salakhutdinov, R.; and Goh, H. 2021.
\newblock Uncertainty weighted actor-critic for offline reinforcement learning.
\newblock \emph{arXiv preprint arXiv:2105.08140}.

\bibitem[{Xiao, Yan, and Amit(2020)}]{OODVAE3}
Xiao, Z.; Yan, Q.; and Amit, Y. 2020.
\newblock Likelihood regret: An out-of-distribution detection score for variational auto-encoder.
\newblock \emph{Advances in neural information processing systems}, 33: 20685--20696.

\bibitem[{Yoshihashi et~al.(2019)Yoshihashi, Shao, Kawakami, You, Iida, and Naemura}]{OODVAE11}
Yoshihashi, R.; Shao, W.; Kawakami, R.; You, S.; Iida, M.; and Naemura, T. 2019.
\newblock Classification-reconstruction learning for open-set recognition.
\newblock In \emph{Proceedings of the IEEE/CVF Conference on Computer Vision and Pattern Recognition}, 4016--4025.

\bibitem[{Zhang et~al.(2023)Zhang, Shao, He, Jiang, and Ji}]{uncer3}
Zhang, H.; Shao, J.; He, S.; Jiang, Y.; and Ji, X. 2023.
\newblock DARL: Distance-Aware Uncertainty Estimation for Offline Reinforcement Learning.
\newblock In \emph{Proceedings of the AAAI Conference on Artificial Intelligence}, volume~37, 11210--11218.

\bibitem[{Zhou(2022)}]{OODVAE9}
Zhou, Y. 2022.
\newblock Rethinking reconstruction autoencoder-based out-of-distribution detection.
\newblock In \emph{Proceedings of the IEEE/CVF Conference on Computer Vision and Pattern Recognition}, 7379--7387.

\end{thebibliography}

\end{document}